\documentclass[runningheads]{llncs}
\usepackage{graphicx}
\usepackage{comment}
\usepackage{amsmath,amssymb} % define this before the line numbering.
\usepackage{color}

\usepackage[utf8]{inputenc} % allow utf-8 input
\usepackage[T1]{fontenc}    % use 8-bit T1 fonts
\usepackage{url}            % simple URL typesetting
\usepackage{booktabs}       % professional-quality tables
\usepackage{nicefrac}       % compact symbols for 1/2, etc.
\usepackage{microtype}      % microtypography
\usepackage{amsmath}
\usepackage[dvipsnames]{xcolor}
\usepackage{mathtools} 
\DeclareMathOperator*{\argmax}{arg\,max}

\usepackage{courier}
\usepackage{color, colortbl}
\definecolor{Gray}{gray}{0.95}

\usepackage[pagebackref=true,breaklinks=true,colorlinks,bookmarks=false]{hyperref}
\usepackage{breakcites}

\usepackage{multirow}
\usepackage{graphicx}  
\usepackage{amsfonts}
\usepackage{bbm}
\usepackage{pifont}  
\newcommand{\cmark}{\ding{51}}

\usepackage[font=small,labelfont=bf]{caption}
\usepackage[caption=false]{subfig}

\newcommand{\norm}[1]{\left\lVert#1\right\rVert}

\newcommand*{\affmark}[1][*]{\textsuperscript{#1}}
\usepackage{wrapfig}
\usepackage[symbol]{footmisc}

\begin{document}
\pagestyle{headings}
\mainmatter
\def\ECCVSubNumber{2343}
\def\eg{\emph{e.g.}}
\def\ie{\emph{i.e.}}
\def\etal{\emph{et al.} }

\title{Learning to Compose Hypercolumns \\for Visual  Correspondence}

\titlerunning{Learning to Compose Hypercolumns for Visual  Correspondence}

\author{Juhong Min\affmark[1,2]\hspace{0.4cm}
Jongmin Lee\affmark[1,2]\hspace{0.4cm}
Jean Ponce\affmark[3,4]\hspace{0.4cm}
Minsu Cho\affmark[1,2]
}

\authorrunning{Juhong Min, Jongmin Lee, Jean Ponce, and Minsu Cho}

\institute{\affmark[1]POSTECH\footnotemark[1]\hspace{0.88cm}
\affmark[2]NPRC\footnotemark[2] \hspace{0.8cm}
\affmark[3]Inria\hspace{0.8cm}
\affmark[4]ENS\footnotemark[3]\\
{\tt\small \url{http://cvlab.postech.ac.kr/research/DHPF/}}}
\maketitle

% !TEX root = ../main.tex

%%%%%%%%% 0. ABSTRACT
\begin{abstract}
Feature representation plays a crucial role in visual correspondence, and recent methods for image matching resort to deeply stacked convolutional layers.  
These models, however, are both monolithic and static in the sense that they typically use a specific level of features, \eg, the output of the last layer, and adhere to it regardless of the images to match. 
In this work, we introduce a novel approach to visual correspondence that dynamically composes effective features by leveraging relevant layers conditioned on the images to match.  
Inspired by both multi-layer feature composition in object detection and adaptive inference architectures in classification, the proposed method, dubbed {\em Dynamic Hyperpixel Flow}, learns to compose hypercolumn features on the fly by selecting a small number of relevant layers from a deep convolutional neural network. 
We demonstrate the effectiveness on the task of semantic correspondence, \ie, establishing correspondences between images depicting different instances of the same object or scene category. 
Experiments on standard benchmarks show that the proposed method greatly improves matching performance over the state of the art in an adaptive and efficient manner. 
\end{abstract}

\footnotetext[1]{Pohang University of Science and Technology, Pohang, Korea}
\footnotetext[2]{The Neural Processing Research Center, Seoul, Korea}
\footnotetext[3]{\'Ecole normale sup\'erieure, CNRS, PSL Research University, 75005 Paris, France}

% !TEX root = ../main.tex

%%%%%%%%% 1. INTRODUCTION
\section{Introduction}

Visual correspondence is at the heart of image understanding with numerous applications such as object recognition, image retrieval, and 3D reconstruction~\cite{forsyth:hal-01063327}. 
With recent advances in neural networks \cite{he2016deep,hu2017senet,huang2015dense,krizhevsky2012imagenet,simonyan2015vgg}, there has been a significant progress in learning robust feature representation for establishing correspondences between images under illumination and viewpoint changes. 
Currently, the de facto standard is to use as feature representation the output of deeply stacked convolutional layers in a trainable architecture.
Unlike in object classification and detection, however, such learned features have often achieved only modest performance gains over hand-crafted
ones~\cite{dalal2005histograms,lowe2004sift} in the task of visual correspondence~\cite{schonberger2017comparative}. 
In particular, correspondence between images under large intra-class variations still remains an extremely challenging problem~\cite{choy2016universal,fathy2018hierarchical,han2017scnet,jeon2018parn,kanazawa2016warpnet,kim2018recurrent,kim2017fcss,kim2017dctm,lee2019sfnet,long2014convnets,min2019hyperpixel,novotny2017anchornet,rocco17geocnn,rocco18weak,rocco2018neighbourhood,paul2018attentive,ufer2017deep,zhou2016learning} while modern neural networks are known to excel at classification~\cite{he2016deep,huang2015dense}.   
What do we miss in using deep neural features for correspondence? 

Most current approaches for correspondence build on monolithic and static feature representations 
in the sense that they use a specific feature layer, \eg, the last convolutional layer, and adhere to it regardless of the images to match. 
Correspondence, however, is all about precise localization of corresponding positions, which requires visual features at different levels, from local patterns to semantics and context; in order to disambiguate a match on similar patterns, it is necessary to analyze finer details and larger context in the image. 
Furthermore, relevant feature levels may vary with the images to match; the more we already know about images, the better we can decide which levels to use. In this aspect, conventional feature representations have fundamental limitations.

In this work, we introduce a novel approach to visual correspondence that dynamically composes effective features by leveraging relevant layers conditioned on the images to match.  
Inspired by both multi-layer feature composition, \ie, hypercolumn, in object detection~\cite{hariharan2015hypercolumns,kong2016hypernet,lin2017feature,liu2018receptive} and adaptive inference architectures in classification~\cite{figurnov2017spatially,srivastava2015highway,veit2018convolutional}, we combine the best of both worlds for visual correspondence.
The proposed method learns to compose hypercolumn features on the fly by selecting a small number of relevant layers in a deep convolutional neural network.   
At inference time, this dynamic architecture greatly improves matching performance in an adaptive and efficient manner.
We demonstrate the effectiveness of the proposed method on several benchmarks for semantic correspondence, \ie, establishing visual correspondences between images depicting different instances of the same object or scene categories, where due to large variations it may be crucial to use features at different levels.

% !TEX root = ../main.tex

%%%%%%%%% 2. RELATED WORK
\section{Related work}

\noindent \textbf{Feature representation for semantic correspondence.}
Early approaches~\cite{bristow2015dense,cho2015unsupervised,ham2016proposal,kim2013deformable,liu2011sift,taniai2016joint,yang2017object} tackle the problem of visual correspondence using hand-crafted descriptors such as HOG~\cite{dalal2005histograms} and SIFT~\cite{lowe2004sift}. 
Since these lack high-level image semantics, the corresponding methods have difficulties with significant changes in background, view point, deformations, and instance-specific patterns.
The advent of convolutional neural networks (CNN)~\cite{he2016deep,krizhevsky2012imagenet} has led to a paradigm shift from this hand-crafted representations to deep features and boosted performance in visual correspondence~\cite{fathy2018hierarchical,novotny2017anchornet,zhou2016learning}. 
Most approaches~\cite{choy2016universal,han2017scnet,kim2017fcss,rocco2018neighbourhood} learn to predict correlation scores between local regions in an input image pair, and some recent methods~\cite{jeon2018parn,kanazawa2016warpnet,kim2018recurrent,rocco17geocnn,rocco18weak,paul2018attentive} cast this task as an image alignment problem in which a model learns to regress global geometric transformation parameters.
All typically adopt a CNN pretrained on image classification as their backbone, and make predictions based on features from its final convolutional layer. 
While some methods~\cite{long2014convnets,zeiler2014visual} have demonstrated the advantage of using different CNN layers in capturing low-level to high-level patterns, leveraging multiple layers of deeply stacked layers has remained largely unexplored in correspondence problems.

\smallbreak
\noindent \textbf{Multi-layer neural features.}
To capture different levels of information distributed over all intermediate layers, Hariharan \etal propose the hypercolumn~\cite{hariharan2015hypercolumns}, a vector of multiple intermediate convolutional activations lying above a pixel for fine-grained localization. Attempts at integrating multi-level neural features have addressed object detection and segmentation~\cite{kong2016hypernet,lin2017feature,liu2018receptive}. In the area of visual correspondence, only a few methods~\cite{min2019hyperpixel,novotny2017anchornet,ufer2017deep} attempt to use multi-layer features. Unlike ours, however, these models use static features extracted from CNN layers that are chosen manually~\cite{novotny2017anchornet,ufer2017deep} or by greedy search~\cite{min2019hyperpixel}. While the use of hypercolumn features on the task of semantic visual correspondence has recently been explored by Min~\etal\cite{min2019hyperpixel}, the method predefines hypercolumn layers by a greedy selection procedure, \ie, beam search, using a validation dataset. 
In this work, we clearly demonstrate the benefit of a dynamic and learnable architecture both in strongly-supervised and weakly-supervised regimes and also outperform the work of~\cite{min2019hyperpixel} with a significant margin.  

\smallbreak
\noindent \textbf{Dynamic neural architectures.}
Recently, dynamic neural architectures have been explored in different domains.
In visual question answering, neural module networks~\cite{andreas2016learning,andreas2016neural} compose different answering networks conditioned on an input sentence. 
In image classification, adaptive inference networks~\cite{figurnov2017spatially,srivastava2015highway,veit2018convolutional} learn to decide whether to execute or bypass intermediate layers given an input image.
Dynamic channel pruning methods~\cite{gao2018dynamic,hua2018channel} skip unimportant channels at run-time to accelerate inference. All these methods reveal the benefit of dynamic neural architectures in terms of either accuracy or speed, or both. 
To the best of our knowledge, our work is the first that explores a dynamic neural architecture for visual correspondence.

Our main contribution is threefold: (1) We introduce a novel dynamic feature composition approach to visual correspondence that composes features on the fly by selecting relevant layers conditioned on images to match. (2) We propose a trainable layer selection architecture for hypercolumn composition using Gumbel-softmax feature gating. (3) The proposed method outperforms recent state-of-the-art methods on standard benchmarks of semantic correspondence in terms of both accuracy and speed. 
% !TEX root = ../main.tex

%%%%%%%%% 3. METHOD
\section{Dynamic hyperpixel flow}
\label{methods}
\begin{figure*}[t]
    \begin{center}

    \scalebox{0.41}{
    \centering
    \includegraphics{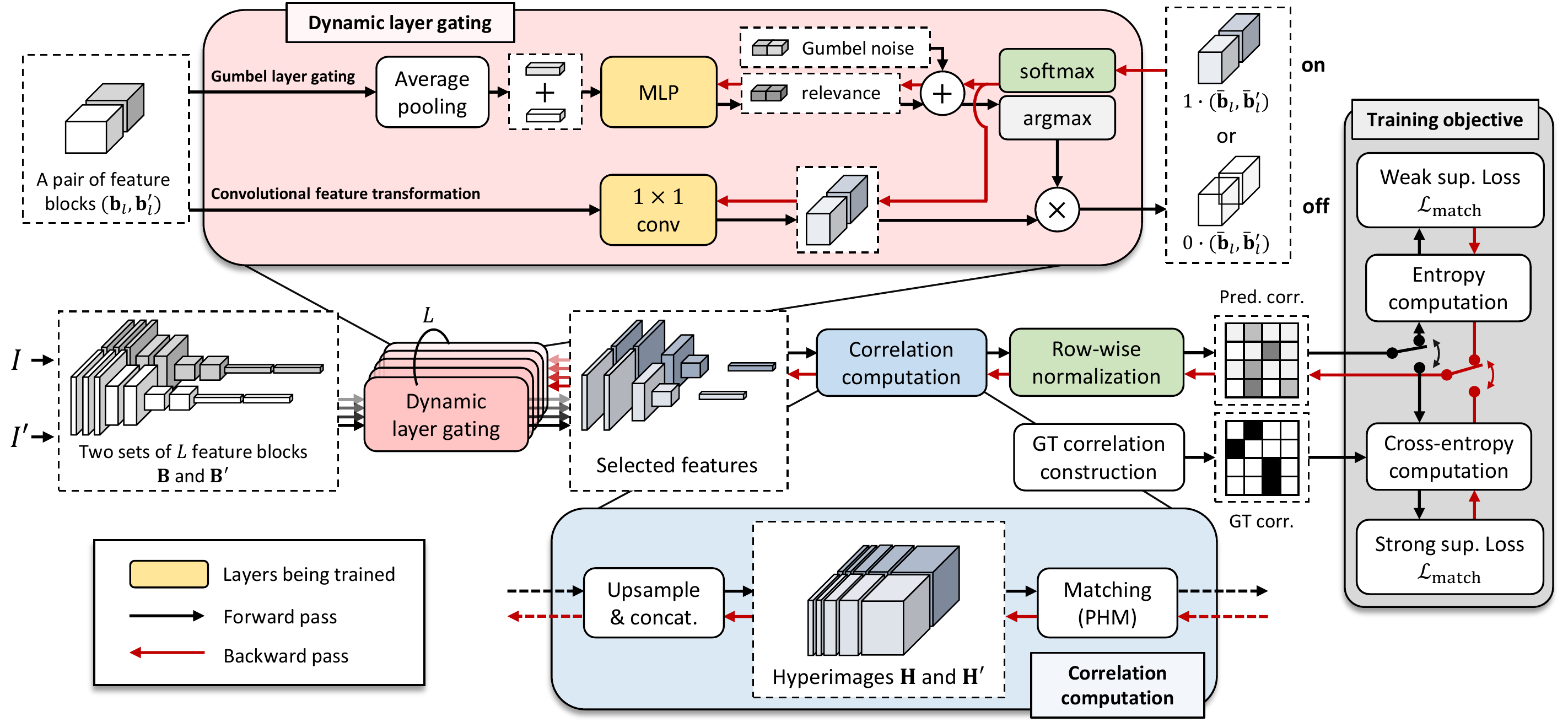}
    }
    \caption{The overall architecture of Dynamic Hyperpixel Flow (DHPF).}
    \label{fig:architecture}
    \end{center}    
\end{figure*}

Given two input images to match, a pretrained convolutional network extracts a series of intermediate feature blocks for each image. The architecture we propose in this section, {\em dynamic hyperpixel flow}, learns to select a small number of layers (feature blocks) on the fly and composes effective features for reliable matching of the images. Figure~\ref{fig:architecture} illustrates the overall architecture.  
In this section, we describe the proposed method in four steps: (i) multi-layer feature extraction, (ii) dynamic layer gating, (iii) correlation computation and matching, and (iv) training objective.

\subsection{Multi-layer feature extraction}

We adopt as a feature extractor a convolutional neural network pretrained on a large-scale classification dataset, \eg, ImageNet~\cite{deng2009imagenet}, which is commonly used in most related methods~\cite{choy2016universal,han2017scnet,kim2018recurrent,kim2017dctm,lee2019sfnet,min2019hyperpixel,rocco17geocnn,rocco18weak,rocco2018neighbourhood,paul2018attentive,huang2019dynamic}. 
Following the work on hypercolumns~\cite{hariharan2015hypercolumns}, however, we view the layers of the convolutional network as a non-linear counterpart of image pyramids and extract a series of multiple features along intermediate layers~\cite{min2019hyperpixel}.  

Let us assume the backbone network contains $L$ feature extracting layers. 
Given two images $I$ and $I'$, source and target, the network generates two sets of $L$ intermediate feature blocks. 
We denote the two sets of feature blocks by $\mathbf{B} = \{\mathbf{b}_l \}_{l=0}^{L-1}$ and  $\mathbf{B}' = \{\mathbf{b}'_l\}_{l=0}^{L-1}$, respectively, and call the earliest blocks, $\mathbf{b}_0$ and $\mathbf{b}'_0$, {\em base} feature blocks.
As in Fig.~\ref{fig:architecture}, each pair of source and target feature blocks at layer $l$ is passed to the $l$-th layer gating module as explained next. 

\subsection{Dynamic layer gating}
Given $L$ feature block pairs  $\{ (\mathbf{b}_l, \mathbf{b}'_l) \}_{l=0}^{L-1}$, $L$ layer gating modules learn to select relevant feature block pairs and transform them for establishing robust correspondences.
As shown in the top of Fig.~\ref{fig:architecture}, the module has two branches, one for layer gating and the other for feature transformation. 

\smallbreak
\noindent \textbf{Gumbel layer gating.} 
The first branch of the $l$-th layer gating module takes the $l$-th pair of feature blocks $(\mathbf{b}_l, \mathbf{b}'_l)$ as an input and performs global average pooling on two feature blocks to capture their channel-wise statistics. Two average pooled features of size $1 \times 1 \times c_l$ from $\mathbf{b}_l$ and  $\mathbf{b}_l' $  are then added together to form a vector of size $c_l$. A multi-layer perceptron (MLP) composed of two fully-connected layers with ReLU non-linearity takes the vector and predicts a relevance vector $\mathbf{r}_l$ of size 2 for gating, whose entries indicate the scores for selecting  or skipping (`on' or `off')  the $l$-th layer, respectively. We can simply obtain a gating decision using argmax over the entries, but this  na\"ive gating precludes backpropagation since argmax is not differentiable. 

To make the layer gating trainable and effective, we adopt the Gumbel-max trick~\cite{gumbel1954statistical} and its continuous relaxation~\cite{eric2017categorical,maddison2017concrete}.
Let $\mathbf{z}$ be a sequence of i.i.d. Gumbel random noise and let $Y$ be a discrete random variable with $K$-class categorical distribution $\mathbf{u}$, \ie, $p(Y=y) \propto u_y$ and $y \in \{0,...,K-1\}$. 
Using the Gumbel-max trick~\cite{gumbel1954statistical}, we can reparamaterize sampling $Y$ to $y = \argmax_{k \in \{0,...,K-1\}}(\log u_k + z_k)$.  
To approximate the argmax in a differentiable manner, the continuous relaxation~\cite{eric2017categorical,maddison2017concrete} of the Gumbel-max trick replaces the argmax operation with a softmax operation. 
By expressing a discrete random sample $y$ as a one-hot vector $
\mathbf{y}$, a sample from the Gumbel-softmax can be represented by $\mathbf{\hat{y}} = \text{softmax}((\log \mathbf{u} + \mathbf{z})/\tau)$, where $\tau$ denotes the temperature of the softmax. 
In our context, the discrete random variable obeys a Bernoulli distribution, \ie, $y \in \{0,1\}$, and the predicted relevance scores represent the log probability distribution for `on' and `off', \ie, $\log{\mathbf{u}} = \mathbf{r}_l$. Our Gumbel-softmax gate thus has a form of 
\begin{align}
\mathbf{\hat{y}}_l = \text{softmax}(\mathbf{r}_l + \mathbf{z}_l), \label{eq:gumbel-softmax}
\end{align}
where $\mathbf{z}_l$ is a pair of i.i.d. Gumbel random samples and the softmax temperature $\tau$ is set to 1. 

\smallbreak
\noindent \textbf{Convolutional feature transformation.} The second branch of the $l$-th layer gating module takes the $l$-th pair of feature blocks $(\mathbf{b}_l, \mathbf{b}'_l)$ as an input and transforms each feature vector over all spatial positions while reducing its dimension by $\frac{1}{\rho}$; we implement it using $1\times1$ convolutions, \ie, position-wise linear transformations, followed by ReLU non-linearity. 
This branch is designed to transform the original feature block of size $h_l \times w_l \times c_l$  into a more compact and effective representation of size $h_l \times w_l \times \frac{c_l}{\rho}$ for our training objective. We denote the pair of transformed feature blocks by $(\bar{\mathbf{b}_l}, \bar{\mathbf{b}'_l})$. 
Note that if $l$-th Gumbel gate chooses to skip the layer, then the feature transformation of the layer can be also ignored thus reducing the computational cost.

\smallbreak
\noindent \textbf{Forward and backward propagations.} During training, we use the {\em straight-through} version of the Gumbel-softmax estimator~\cite{eric2017categorical}: forward passes proceed with discrete samples by argmax whereas backward passes compute gradients of the softmax relaxation of Eq.(\ref{eq:gumbel-softmax}). In the forward pass, the transformed feature pair $(\bar{\mathbf{b}_l}, \bar{\mathbf{b}'_l})$ is simply multiplied by 1 (`on') or 0 (`off') according to the gate's discrete decision $\mathbf{y}$. 
While the Gumbel gate always makes discrete decision $\mathbf{y}$ in the forward pass, the continuous relaxation in the backward pass allows gradients to propagate through softmax output $\hat{\mathbf{y}}$, effectively updating both branches, the feature transformation and the relevance estimation,  regardless of the gate's decision.
Note that this stochastic gate with random noise increases the diversity of samples and is thus crucial in preventing mode collapse in training. At test time, we simply use deterministic gating by argmax without Gumbel noise~\cite{eric2017categorical}.
As discussed in Sec.~\ref{sec:softgating}, we found that the proposed hard gating trained with Gumbel softmax is superior to conventional soft gating with sigmoid in terms of both accuracy and speed.

\subsection{Correlation computation and matching}
The output of gating is a set of selected layer indices, $S = \{s_1, s_2, ..., s_N\}$. We construct a {\em hyperimage} $\mathbf{H}$ for each image by concatenating transformed feature blocks of the selected layers along channels with upsampling: 
$\mathbf{H} = \big[ \zeta(\bar{\mathbf{b}_{s_1}}), \zeta(\bar{\mathbf{b}_{s_2}}), ..., \zeta(\bar{\mathbf{b}_{s_N}}) \big]$,
where $\zeta$ denotes a function that spatially upsamples the input feature block to the size of $\mathbf{b}_0$, the {\em base} block. Note that the number of selected layers $N$ is fully determined by the gating modules. If all layers are off, then we use the base feature block by setting $S = \{0\}$. We associate with each spatial position $p$ of the hyperimage the corresponding image coordinates and hyperpixel feature~\cite{min2019hyperpixel}. Let us denote by $\mathbf{x}_p$ the image coordinate of position $p$, and by  $\mathbf{f}_p$ the corresponding feature, {\em \ie}, $\mathbf{f}_p = \mathbf{H}({\mathbf{x}_p})$.
The hyperpixel at position $p$ in the hyperimage is defined as
  $ \mathbf{h}_p = (\mathbf{x}_p, \mathbf{f}_p)$. Given source and target images, we obtain two sets of hyperpixels, $\mathcal{H}$ and $\mathcal{H}'$. 
In order to reflect geometric consistency in matching, we adapt probablistic Hough matching (PHM)~\cite{cho2015unsupervised,han2017scnet} to hyperpixels, similar to~\cite{min2019hyperpixel}. The key idea of PHM is to re-weight appearance similarity by Hough space voting to enforce geometric consistency. In our context, let $\mathcal{D}=(\mathcal{H}, \mathcal{H}')$ be two sets of hyperpixels, and $m=(\mathbf{h},\mathbf{h}')$ be a match where $\mathbf{h}$ and $\mathbf{h}'$ are respectively elements of $\mathcal{H}$ and $\mathcal{H}'$. Given a Hough space $\mathcal{X}$ of possible offsets (image transformations) between the two hyperpixels, the confidence for match $m$, $p(m|\mathcal{D})$, is computed as  
$p(m|\mathcal{D}) \propto  p(m_\mathrm{a})\sum_{\mathbf{x}\in \mathcal{X}}p(m_\mathrm{g}|\mathbf{x})\sum_{m \in \mathcal{H} \times \mathcal{H}'}p(m_\mathrm{a})p(m_\mathrm{g}|\mathbf{x})$ 
where $p(m_\mathrm{a})$ represents the confidence for appearance matching and $p(m_\mathrm{g}|\mathbf{x})$ is the confidence for geometric matching with an offset $\mathbf{x}$, measuring how close the offset induced by $m$ is to $\mathbf{x}$. By sharing the Hough space $\mathcal{X}$ for all matches, PHM efficiently computes match confidence with good empirical performance~\cite{cho2015unsupervised,ham2016proposal,han2017scnet,min2019hyperpixel}. In this work, we compute appearance matching confidence using hyperpixel features by 
$p(m_\mathrm{a}) \propto \text{ReLU}\Big( \frac{\mathbf{f}_p \cdot \mathbf{f}_p'}{\norm{\mathbf{f}_p} \norm{\mathbf{f}_p'}} \Big)^2,$
where the squaring has the effect of suppressing smaller matching confidences. 
On the output $|\mathcal{H}| \times |\mathcal{H}'|$ correlation matrix of PHM, we perform soft mutual nearest neighbor filtering~\cite{rocco2018neighbourhood} to suppress noisy correlation values and denote the filtered matrix by $\mathbf{C}$.

\smallbreak
\noindent \textbf{Dense matching and keypoint transfer.} 
From the correlation matrix $\mathbf{C}$, we establish hyperpixel correspondences by assigning to each source hyperpixel $\mathbf{h}_i$ the target hyperpixel $\hat{\mathbf{h}}'_{j}$ with the highest correlation. Since the spatial resolutions of the hyperimages are the same as those of base feature blocks, which are relatively high in most cases (\eg, $1/4$ of input image with ResNet-101 as the backbone), such hyperpixel correspondences produce quasi-dense matches.

Furthermore, given a keypoint $\mathbf{p}_m$ in the source image, we can easily predict its corresponding position $\hat{\mathbf{p}}'_m$ in the target image by transferring the keypoint using its nearest hyperpixel correspondence. In our experiments, we collect all correspondences of neighbor hyperpixels of keypoint $\mathbf{p}_m$ and use the geometric average of their individual transfers as the final prediction $\hat{\mathbf{p}}'_m$~\cite{min2019hyperpixel}. 
This consensus keypoint transfer method improves accuracy by refining mis-localized predictions of individual transfers.

\begin{figure}[!tbp]
  \centering
  \begin{minipage}[b]{0.49\textwidth}
    \subfloat[Strongly-supervised loss.]{
        \includegraphics[width=\textwidth]{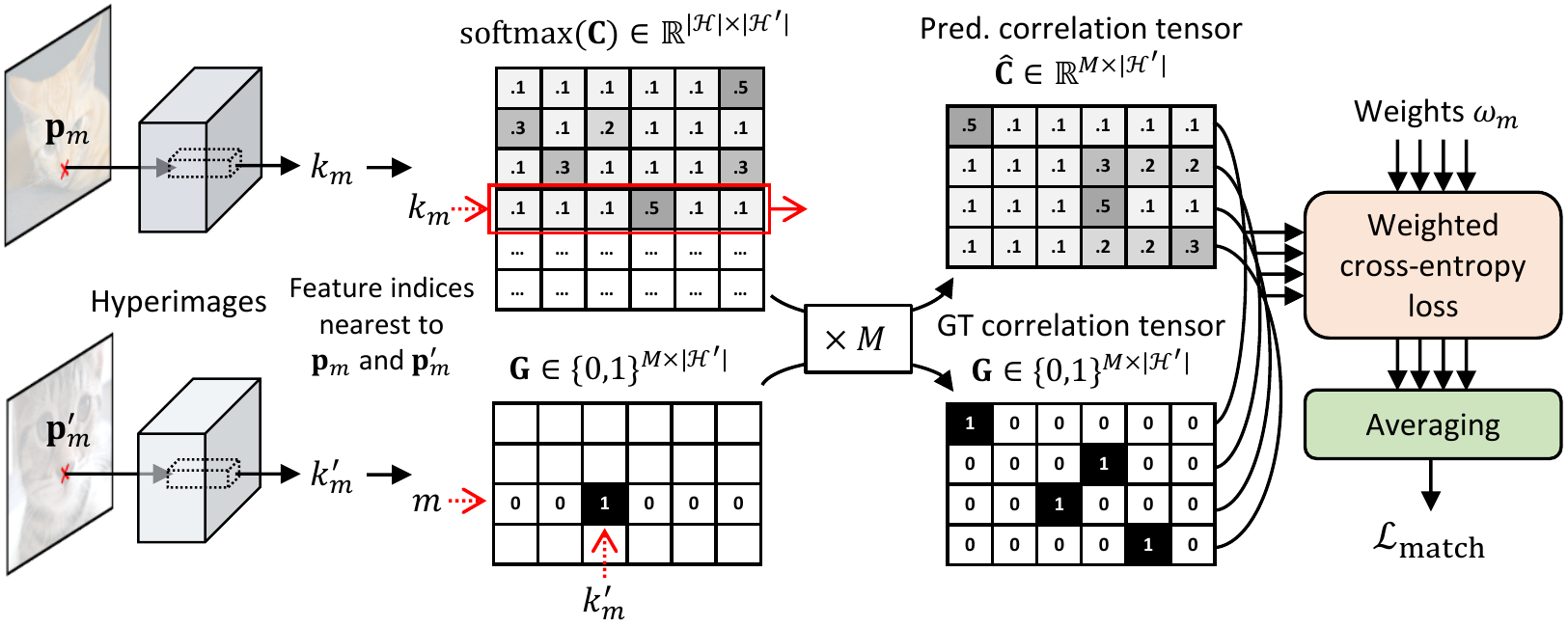}
        }
  \end{minipage}
  \hfill
  \begin{minipage}[b]{0.49\textwidth}
    \subfloat[Weakly-supervised loss.]{
        \includegraphics[width=\textwidth]{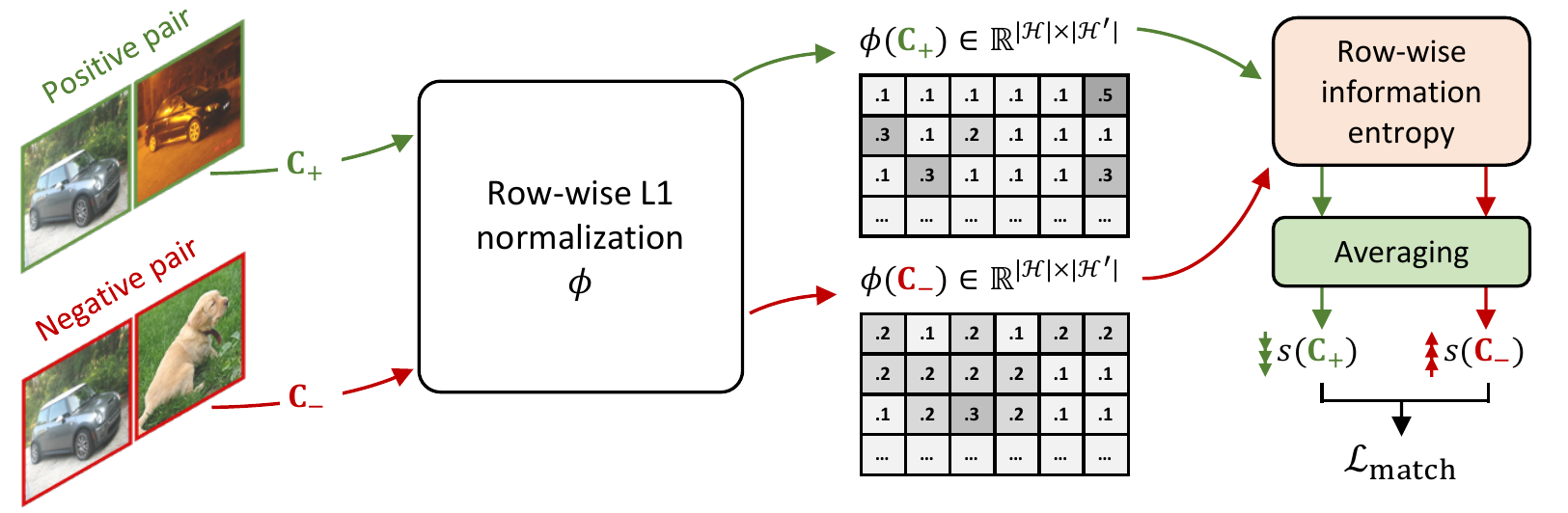}
        }
  \end{minipage}
  
    \caption{Matching loss computation using (a) keypoint annotations (strong supervision) and (b) image pairs only (weak supervision). Best viewed in electronic form.}
    \label{fig:loss_computation}
\end{figure}

\subsection{Training objective}
We propose two objectives to train our model using different degrees of supervision: strongly-supervised and weakly-supervised regimes.

\smallbreak
\noindent \textbf{Learning with strong supervision.} 
In this setup, we assume that keypoint match annotations are given for each training image pair, as in~\cite{choy2016universal,han2017scnet,min2019hyperpixel}; each image pair is annotated with a set of coordinate pairs $\mathcal{M}=\{(\mathbf{p}_m, \mathbf{p}'_m)\}_{m=1}^{M}$, where $M$ is the number of match annotations.  

To compare the output of our network with ground-truth annotations, we convert the annotations into a form of discrete correlation matrix. 
First of all, for each coordinate pair $(\mathbf{p}_m, \mathbf{p}'_m)$, we identify their nearest position indices $(k_m, k_m')$ in hyperimages. 
On the one hand, given the set of identified match index pairs $\{ (k_m, k_m') \}_{m=1}^{M}$, we construct a ground-truth matrix $\mathbf{G} \in \{0,1\}^{M\times |\mathcal{H}'|}$ by assigning one-hot vector representation of $k_m'$ to the $m$-th row of  $\mathbf{G}$. 
On the other hand, we construct $\hat{\mathbf{C}} \in \mathbb{R}^{M \times |\mathcal{H}'|}$ by assigning the $k_m$-th row of $\mathbf{C}$ to the $m$-th row of $\hat{\mathbf{C}}$. 
We apply softmax to each row of the matrix $\hat{\mathbf{C}}$ after normalizing it to have zero mean and unit variance.
Figure~\ref{fig:loss_computation}a illustrates the construction of $\hat{\mathbf{C}}$ and $\mathbf{G}$. 
Corresponding rows between $\hat{\mathbf{C}}$ and $\mathbf{G}$ can now be compared as categorical probability distributions. 
We thus define the strongly-supervised matching loss as the sum of cross-entropy values between them: 
\begin{align}
     \label{one_hot_gce}
     \mathcal{L}_\mathrm{match} = - \frac{1}{M} \sum_{m=1}^{ M} \omega_{m} \sum_{j = 1}^{|\mathcal{H}'|}\mathbf{G}_{m j}\log\hat{\mathbf{C}}_{mj}, 
\end{align}
where $\omega_{m}$ is an importance weight for the $m$-th keypoint. The keypoint weight $\omega_{m}$ helps training by reducing the effect of the corresponding cross-entropy term if the Eucliean distance between predicted keypoint $\hat{\mathbf{p}}'_m$ and target keypoint $\mathbf{p'}_m$ is smaller than some threshold distance $\delta_\mathrm{thres}$: 
\begin{align}
     \label{weighting_term}
    \omega_{m} = 
    \begin{dcases}
        (\norm{\hat{\mathbf{p}}'_m-\mathbf{p}'_m} / \delta_{\mathrm{thres}})^2 & \text{if} \quad \norm{\hat{\mathbf{p}}'_m-\mathbf{p}'_m} < \delta_{\mathrm{thres}}, \\
        1 & \text{otherwise.}
    \end{dcases}
\end{align}

The proposed objective for strongly-supervised learning can also be used for self-supervised learning with synthetic  pairs~\cite{rocco17geocnn,paul2018attentive}\footnote{For example, we can obtain keypoint annotations for free by forming a synthetic pair by applying random geometric transformation (\eg, affine or TPS~\cite{donato2002approximate}) on an image and then sampling some corresponding points between the original image and the warped image using the transformation applied.}, which typically results in trading off the cost of supervision against the generalization performance.

\smallbreak
\noindent \textbf{Learning with weak supervision.} In this setup, we assume that only image-level labels are given for each image pair as either positive (the same class) or negative (different class), as in~\cite{huang2019dynamic,rocco2018neighbourhood}.
Let us denote the correlation matrix of a positive pair by  $\mathbf{C_{+}}$ and that of a negative pair by $\mathbf{C_{-}}$.
For $\mathbf{C} \in \mathbb{R}^{|\mathcal{H}| \times |\mathcal{H}'|}$, we define its correlation entropy as 
$s(\mathbf{C}) = -\frac{1}{|\mathcal{H}|}\sum_{i=1}^{|\mathcal{H}|}\sum_{j=1}^{|\mathcal{H}'|}\phi(\mathbf{C})_{ij}\log\phi(\mathbf{C})_{ij}$
where $\phi(\cdot)$ denotes row-wise L1-normalization. Higher correlation entropy indicates less distinctive correspondences between the two images. As illustrated in Fig.~\ref{fig:loss_computation}b, assuming that the positive images are likely to contain more distinctive correspondences, we encourage low entropy for positive pairs and high entropy for negative pairs.
The weakly-supervised matching loss is formulated as 
\begin{align}
     \mathcal{L}_\mathrm{match} = \frac{s(\mathbf{C_{+}}) + s(\mathbf{C_{+}^{\top}})} {s(\mathbf{C_{-}}) + s(\mathbf{C_{-}^{\top}})}.
\end{align}

\smallbreak
\noindent \textbf{Layer selection loss.} 
Following the work of~\cite{veit2018convolutional}, we add a soft constraint in our training objective to encourage the network to select each layer at a certain rate: 
$\mathcal{L}_\mathrm{sel} = \sum_{l=0}^{L-1} (\Bar{z}_l - \mu)^2$
where $\Bar{z}_l$ is a fraction of image pairs within a mini-batch for which the $l$-th layer is selected and $\mu$ is a hyperparameter for the selection rate.
This improves training by increasing diversity in layer selection and, as will be seen in our experiments, allows us to trade off between accuracy and speed in testing.    

Finally, the training objective of our model is defined as the combination of the matching loss (either strong or weak) and the layer selection loss: $\mathcal{L} = \mathcal{L_{\mathrm{match}}} + \mathcal{L_{\mathrm{sel}}}$.

% !TEX root = ../main.tex

%%%%%%%%% 4. EXPERIMENT
\begin{table}[!t]
    \caption{\label{tab:main_table}Performance on standard benchmarks in accuracy and speed (avg. time per pair). The subscript of each method name denotes its feature extractor. Some results are from ~\cite{jeon2018parn,kim2018recurrent,lee2019sfnet,min2019hyperpixel}. Numbers in bold indicate the best performance and underlined ones are the second best. The average inference time (the last column) is measured on test split of PF-PASCAL~\cite{ham2018proposal} and includes all the pipelines of the models: from feature extraction to keypoint prediction.
    }
    
    \begin{center}
    
    \scalebox{0.83}{
    \begin{tabular}{cclcccccccccc}
            \toprule
            \multirow{3}{*}{Sup.} & \multirow{3}{*}{Sup. signal} & \multirow{3}{*}{Methods} & \multicolumn{4}{c}{PF-PASCAL} & \multicolumn{3}{c}{PF-WILLOW} & \multicolumn{2}{c}{Caltech-101} & time \\
             & & & \multicolumn{3}{c}{PCK @ $\alpha_{\text{img}}$} & $\alpha_{\text{bbox}}$ & \multicolumn{3}{c}{PCK @ $\alpha_{\text{bbox}}$} & \multirow{2}{*}{LT-ACC} & \multirow{2}{*}{IoU} & \multirow{2}{*}{({\em ms})} \\ 
             & & & 0.05 & 0.1 & 0.15 & 0.1 & 0.05 & 0.1 & 0.15 & & & \\
             \midrule
             none & - & PF$_\textrm{HOG}$~\cite{ham2016proposal} & 31.4 & 62.5 & 79.5 & 45.0 & 28.4 & 56.8 & 68.2 & 0.78 & 0.50 & >1000 \\
             \midrule
             \multirow{2}{*}{self} & \multirow{2}{*}{\shortstack{synthetic\\pairs}} & CNNGeo$_\textrm{res101}$~\cite{rocco17geocnn} & 41.0 & 69.5 & 80.4 & 68.0 & 36.9 & 69.2 & 77.8 & 0.79 & 0.56 & 40 \\
             & & A2Net$_\textrm{res101}$~\cite{paul2018attentive}       & 42.8 & 70.8 & 83.3 &  67.0 & 36.3 & 68.8 & 84.4 & 0.80 & 0.57 & 53 \\ 
             \midrule
             \multirow{8}{*}{weak} & \multirow{1}{*}{\shortstack{bbox}} & SF-Net$_\textrm{res101}$~\cite{lee2019sfnet}      & 53.6  & 81.9 & 90.6 & {78.7} & 46.3 & 74.0 & 84.2 & 0.88 & 0.67 & 51  \\\cline{2-13}
             & \multirow{7}{*}{\shortstack{image-level\\labels}} & Weakalign$_\textrm{res101}$~\cite{rocco18weak} & 49.0 & 74.8 & 84.0 & 72.0 & 37.0 & 70.2 & 79.9 & \underline{0.85} & \textbf{0.63} & {41} \\  
             & & RTNs$_\textrm{res101}$~\cite{kim2018recurrent} & 55.2 & 75.9 & 85.2 & - & 41.3 & 71.9 & 86.2 & - & - & 376  \\
             & & NC-Net$_\textrm{res101}$~\cite{rocco2018neighbourhood} & 54.3 & 78.9 & 86.0 & 70.0 & 33.8 & 67.0 & 83.7 & \underline{0.85} & 0.60 & 261 \\
             & & DCC-Net$_\textrm{res101}$~\cite{huang2019dynamic} & \underline{55.6} & \textbf{82.3} & \underline{90.5} & - & 43.6 & 73.8 & 86.5 & - & - & >261 \\\cline{3-13}\\[-2.3ex]
             
             & & DHPF$^{\mu=0.4}_\textrm{res50}$ (ours) & 54.8 & 79.0 & 89.8 & \underline{74.5} & 48.7 & 75.7 & 87.3 & \underline{0.85} & 0.59 & \textbf{31} \\
             
             & & DHPF$_\textrm{res50}$ (ours) & 54.7 & 79.0 & 89.7 & \underline{74.5} & \textbf{51.8} & \underline{78.7} & \underline{89.6} & \underline{0.85} & 0.59 & \underline{33} \\
             
             & & DHPF$_\textrm{res101}$ (ours) & \textbf{56.1} & \underline{82.1} & \textbf{91.1} & \textbf{78.5} & \underline{50.2} & \textbf{80.2} & \textbf{91.1} & \textbf{0.86} & \underline{0.61} & 56 \\
             
             \midrule
             \multirow{6}{*}{strong} & \multirow{6}{*}{\shortstack{src \& trg\\keypoint\\matches}} & SCNet$_\textrm{vgg16}$~\cite{han2017scnet}            & 36.2 & 72.2 & 82.0 & 48.2 & 38.6 & 70.4 & 85.3 & 0.79 & 0.51 &  >1000 \\
             & & HPF$_{\textrm{res50}}$~\cite{min2019hyperpixel}        & 60.5 & 83.4 & 92.1 & 76.5 & 46.5 & 72.4 & 84.7 & \textbf{0.88} & \textbf{0.64} & \underline{34} \\
             & & HPF$_{\textrm{res101}}$~\cite{min2019hyperpixel}       & 60.1 & 84.8 & 92.7 & 78.5 & 45.9 & 74.4 & 85.6 & \underline{0.87} & \underline{0.63} & 63 \\
             \cline{3-13}\\[-2.3ex]
            
             & & DHPF$^{\mu=0.4}_\textrm{res50}$ (ours) & 70.2 & \underline{89.1} & 94.0 & 85.0 & 45.8 & 73.3 & 86.6 & 0.86 & 0.60 & \textbf{30} \\
             
             & & DHPF$_\textrm{res50}$ (ours) & \underline{72.6} & {88.9} & \underline{94.3} & \underline{85.6} & \underline{47.9} & \underline{74.8} & \underline{86.7} & 0.86 & 0.61 & \underline{34} \\
             
             & & DHPF$_\textrm{res101}$ (ours) & \textbf{75.7} &  \textbf{90.7} & \textbf{95.0} & \textbf{87.8} & \textbf{49.5} & \textbf{77.6} & \textbf{89.1} & \underline{0.87} & 0.62 & 58 \\
             
            \bottomrule
    \end{tabular}
    }
    \end{center}
\end{table}

\section{Experiments}

In this section we compare our method to the state of the art and discuss the results. 
The code and the trained model are available online at our project page. 
\smallbreak
\noindent \textbf{Feature extractor networks.} As the backbone networks for feature extraction, we use ResNet-50 and ResNet-101~\cite{he2016deep}, which contains 49 and 100 conv layers in total (excluding the last FC), respectively. Since features from adjacent layers are strongly correlated, we extract the base block from \texttt{conv1} maxpool and intermediate blocks from layers with residual connections (before ReLU). They amounts to 17 and 34 feature blocks (layers) in total, respectively, for ResNet-50 and ResNet-101. Following related work~\cite{choy2016universal,han2017scnet,kim2018recurrent,lee2019sfnet,min2019hyperpixel,rocco17geocnn,rocco18weak,rocco2018neighbourhood,paul2018attentive,huang2019dynamic}, we freeze the backbone network parameters during training for fair comparison.

\smallbreak
\noindent \textbf{Datasets.} Experiments are done on four benchmarks for semantic correspondence: PF-PASCAL~\cite{ham2018proposal}, PF-WILLOW~\cite{ham2016proposal}, Caltech-101~\cite{li2006one}, and SPair-71k~\cite{min2019spair}. PF-PASCAL and PF-WILLOW consist of  keypoint-annotated image pairs, 1,351 pairs from 20 categories, and 900 pairs from 4 categories, respectively. 
Caltech-101~\cite{li2006one} contains  segmentation-annotated 1,515 pairs from 101 categories. SPair-71k~\cite{min2019spair} is a more challenging large-scale dataset recently introduced in~\cite{min2019hyperpixel}, consisting of keypoint-annotated 70,958 image pairs from 18 categories with diverse view-point and scale variations.

\smallbreak
\noindent \textbf{Evaluation metrics.} As an evaluation metric for PF-PASCAL, PF-WILLOW, and SPair-71k, the probability of correct keypoints (PCK) is used. The PCK value given a set of predicted and ground-truth keypoint pairs $\mathcal{P}=\{(\hat{\mathbf{p}}'_{m}, \ \mathbf{p}'_{m})\}_{m=1}^{M}$ is measured by $\mathrm{PCK}(\mathcal{P}) = \frac{1}{M}\sum_{m=1}^{M} \mathbbm{1} [\norm{\hat{\mathbf{p}}'_{m} - \mathbf{p}'_{m}} \leq \alpha_{\tau}  \max{(w_{\tau}, h_{\tau})} ]$.
As an evaluation metric for the Caltech-101 benchmark, the label transfer accuracy (LT-ACC)~\cite{liu2009nonparam} and the  intersection-over-union (IoU)~\cite{Everingham2010} are used. 
Running time (average time per pair) for each method is measured using its authors' code on a machine with an Intel i7-7820X CPU and an NVIDIA Titan-XP GPU.

\smallbreak
\noindent \textbf{Hyperparameters.} The layer selection rate $\mu$ and the channel reduction factor $\rho$ are determined by grid search using the validation split of PF-PASCAL. As a result, we set $\mu=0.5$ and $\rho=8$ in our experiments if not specified otherwise.
The threshold $\delta_{\mathrm{thres}}$ in Eq.(\ref{weighting_term}) is set to be  $\text{max}(w_{\tau}, h_{\tau})/10$.

\begin{table}[!t]
    
    \caption{\label{tab:hpftable} Performance on SPair-71k dataset in accuracy (per-class PCK with $\alpha_{\text{bbox}}=0.1$). TR represents transferred models trained on PF-PASCAL while FT denotes fine-tuned (trained) models on SPair-71k.}
    
    \begin{center}
        \scalebox{0.6}{
        \begin{tabular}{cclccccccccccccccccccc}
        \toprule
         Sup. & \multicolumn{2}{c}{Methods} & aero & bike & bird & boat & bottle & bus & car & cat & chair & cow & dog & horse & mbike & person & plant & sheep & train & tv & all\\
        \midrule
        \multirow{4}{*}{self} & TR
        & CNNGeo$_\textrm{res101}$~\cite{rocco17geocnn} & 21.3 & 15.1 & 34.6 & 12.8 & 31.2 & 26.3 & 24.0 & 30.6 & 11.6 & 24.3 & 20.4 & 12.2 & 19.7 & 15.6 & 14.3 & 9.6 & 28.5 & 28.8 & 18.1  \\
        
        & FT & CNNGeo$_\textrm{res101}$~\cite{rocco17geocnn} &  23.4 & 16.7 & 40.2 & 14.3 & 36.4 & 27.7 & 26.0 & 32.7 & 12.7 & 27.4 & 22.8 & 13.7 & 20.9 & 21.0 & 17.5 & 10.2 & 30.8 & 34.1 & 20.6  \\
        
        & TR & A2Net$_\textrm{res101}$~\cite{paul2018attentive} &  20.8 & 17.1 & 37.4 & 13.9 & 33.6 & {29.4} & {26.5} & 34.9 & 12.0 & 26.5 & 22.5 & 13.3 & 21.3 & 20.0 & 16.9 & 11.5 & 28.9 & 31.6 & 20.1 \\
        
        & FT & A2Net$_\textrm{res101}$~\cite{paul2018attentive} & 22.6 & {18.5} & 42.0 & {16.4} & {37.9} & {30.8} & {26.5} & 35.6 & 13.3 & 29.6 & 24.3 & 16.0 & 21.6 & {22.8} & {20.5} & 13.5 & 31.4 & {36.5} & 22.3 \\
        \midrule
        
        \multirow{6}{*}{weak} & TR & WeakAlign$_\textrm{res101}$~\cite{rocco18weak} & \underline{23.4} & 17.0 & 41.6 & 14.6 & \underline{37.6} & \textbf{28.1} & \underline{26.6} & 32.6 & 12.6 & 27.9 & 23.0 & 13.6 & 21.3 & 22.2 & 17.9 & 10.9 & {31.5} & 34.8 & 21.1 \\
        
        & FT & WeakAlign$_\textrm{res101}$~\cite{rocco18weak} &  22.2 & 17.6 & 41.9 & \underline{15.1} & \textbf{38.1} & \underline{27.4} & \textbf{27.2} & 31.8 & 12.8 & 26.8 & 22.6 & 14.2 & 20.0 & 22.2 & 17.9 & 10.4 & \underline{32.2} & 35.1 & 20.9 \\
        
        & TR & NC-Net$_\textrm{res101}$~\cite{rocco2018neighbourhood} & \textbf{24.0} & 16.0 & {45.0} & 13.7 & 35.7 & 25.9 & 19.0 & \underline{50.4} & {14.3} & {32.6} & {27.4} & \underline{19.2} & {21.7} & 20.3 & 20.4 & {13.6} & \textbf{33.6} & \underline{40.4} & {26.4} \\
        
        & FT & NC-Net$_\textrm{res101}$~\cite{rocco2018neighbourhood} & 17.9 & 12.2 & 32.1 & 11.7 & 29.0 & 19.9 & 16.1 & 39.2 & 9.9 & 23.9 & 18.8 & 15.7 & 17.4 & 15.9 & 14.8 & 9.6 & 24.2 & 31.1 & 20.1   \\\cline{2-22}\\[-2.3ex]

        & TR & {DHPF$_\mathrm{res101}$ (ours)} & 21.5 & \textbf{21.8} & \textbf{57.2} & 13.9 & 34.3 & 23.1 & 17.3 & \underline{50.4} & \textbf{17.4} & \underline{34.8} & \textbf{36.2} & \textbf{19.7} & \underline{24.3} & \textbf{32.5} & \textbf{22.2} & \underline{17.6} & 30.9 & 36.5 & \textbf{28.5} \\
        
        & FT & {DHPF$_\mathrm{res101}$ (ours)} & 17.5 & \underline{19.0} & \underline{52.5} & \textbf{15.4} & 35.0 & 19.4 & 15.7 & \textbf{51.9} & \underline{17.3} & \textbf{37.3} & \underline{35.7} & \textbf{19.7} & \textbf{25.5} & \underline{31.6} & \underline{20.9} & \textbf{18.5} & 24.2 & \textbf{41.1} & \underline{27.7} \\
        
        \midrule

        \multirow{3}{*}{strong} & FT & HPF$_\mathrm{res101}$~\cite{min2019hyperpixel} & \underline{25.2} & {18.9} & {52.1} & \underline{15.7} & \underline{38.0} & \underline{22.8} & \underline{19.1} & \underline{52.9} & {17.9} & \underline{33.0} & {32.8} & \underline{20.6} & \underline{24.4} & {27.9} & 21.1 & \underline{15.9} & \underline{31.5} & \underline{35.6} & \underline{28.2} \\\cline{2-22}\\[-2.3ex]

        & TR & {DHPF$_\mathrm{res101}$ (ours)} & 22.6 & \underline{23.0} & \underline{57.7} & 15.1 & 34.1 & 20.5 & 14.7 & 48.6 & \underline{19.5} & 31.9 & \underline{34.5} & 19.6 & 23.0 & \underline{30.0} & \underline{22.9} & 15.5 & 28.2 & 30.2 & 27.4 \\
        
        & FT & {DHPF$_\mathrm{res101}$ (ours)} & \textbf{38.4} & \textbf{23.8} & \textbf{68.3} & \textbf{18.9} & \textbf{42.6} & \textbf{27.9} & \textbf{20.1} & \textbf{61.6} & \textbf{22.0} & \textbf{46.9} & \textbf{46.1} & \textbf{33.5} & \textbf{27.6} & \textbf{40.1} & \textbf{27.6} & \textbf{28.1} & \textbf{49.5} & \textbf{46.5} & \textbf{37.3} \\
        \bottomrule
        
        \end{tabular}}
    \end{center}
\end{table}

\subsection{Results and comparisons}
First, we train both of our strongly and weakly-supervised models on the PF-PASCAL~\cite{ham2018proposal} dataset and test on three standard benchmarks of PF-PASCAL (test split), PF-WILLOW and Caltech-101. The evaluations on PF-WILLOW and Caltech-101 are to verify transferability.
In training, we use the same splits of PF-PASCAL proposed in~\cite{han2017scnet} where training, validation, and test sets respectively contain 700, 300, and 300 image pairs.  Following \cite{rocco18weak,rocco2018neighbourhood}, we augment the training pairs by horizontal flipping and swapping. Table~\ref{tab:main_table} summarizes our result and those of recent methods~\cite{ham2016proposal,han2017scnet,kim2018recurrent,kim2017dctm,min2019hyperpixel,rocco17geocnn,rocco18weak,rocco2018neighbourhood,paul2018attentive}. 
Second, we train our model on the SPair-71k dataset~\cite{min2019spair} and compare it to other recent methods~\cite{min2019hyperpixel,rocco17geocnn,rocco18weak,rocco2018neighbourhood,paul2018attentive}.
Table~\ref{tab:hpftable} summarizes the results. 

\smallbreak
\noindent \textbf{Strongly-supervised regime.} As shown in the bottom sections of Table~\ref{tab:main_table} and~\ref{tab:hpftable}, our strongly-supervised model clearly outperforms the previous state of the art by a significant margin. It achieves 5.9\%, 3.2\%, and 9.1\% points of PCK ($\alpha_{\text{img}}=0.1$) improvement over the current state of the art~\cite{min2019hyperpixel} on PF-PASCAL, PF-WILLOW, and SPair-71k, respectively, and the improvement increases further with a more strict evaluation threshold, \eg, more than 15\% points of PCK with $\alpha_{\text{img}}=0.05$ on PF-PASCAL. Even with a smaller backbone network (ResNet-50) and smaller selection rate ($\mu = 0.4$), our method achieves competitive performance with the smallest running time on the standard benchmarks of PF-PASCAL, PF-WILLOW, and Caltech-101.

\begin{figure}[!t]
    \begin{center}

    \scalebox{0.33}{
    \centering
    \includegraphics{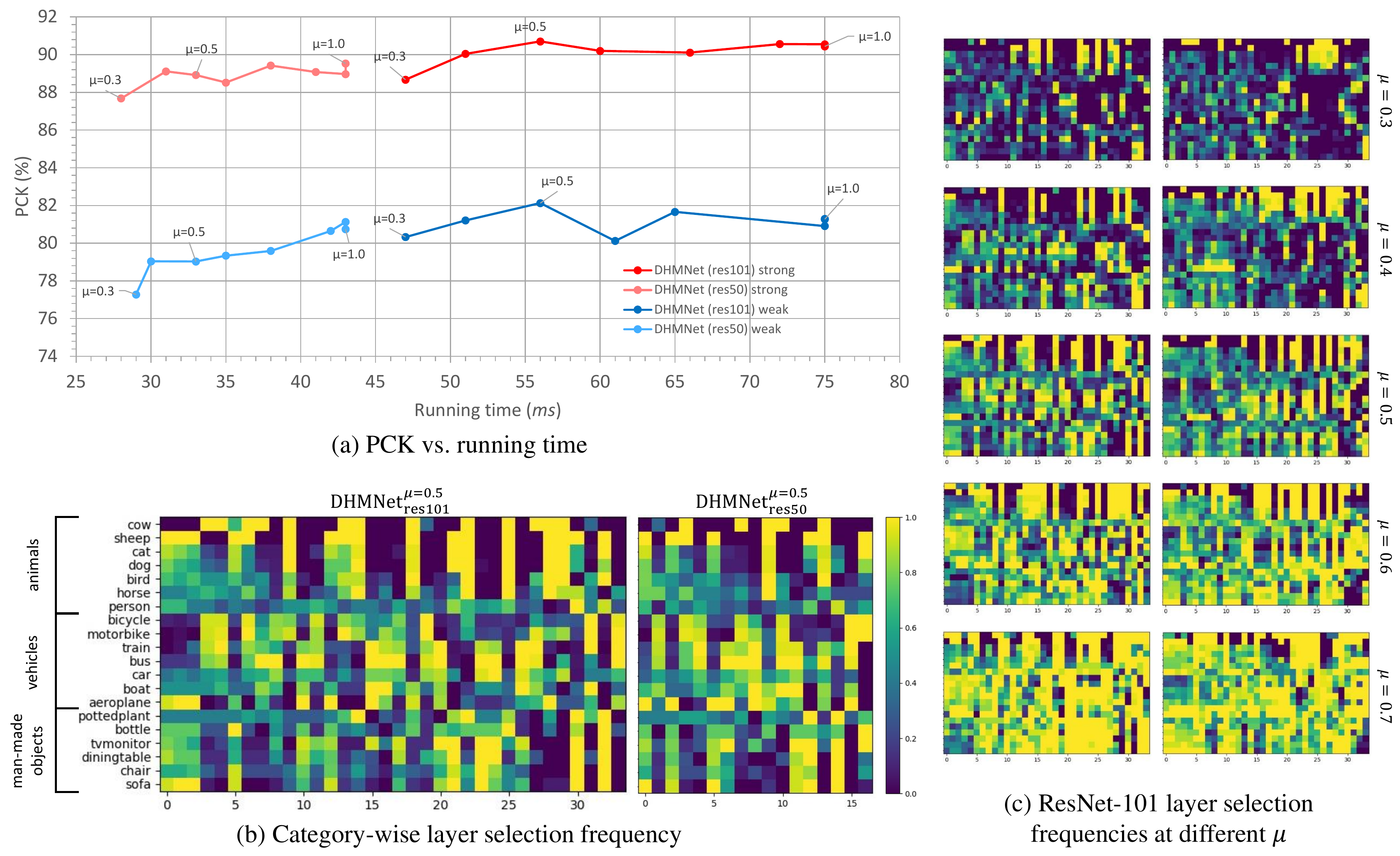}
    }
    \caption{Analysis of layer selection on PF-PASCAL dataset (a) PCK vs. running time with varying selection rate $\mu$ (b) Category-wise layer selection frequencies (x-axis: candidate layer index, y-axis: category) of the strongly-supervised model with different backbones: ResNet-101 (left) and ResNet-50 (right) (c) ResNet-101 layer selection frequencies of strongly (left) and weakly (right) supervised models at different layer selection rates $\mu$. Best viewed in electronic form.}
    \label{fig:layer_selection}
    \end{center}    
\end{figure}

\smallbreak
\noindent \textbf{Weakly-supervised regime.} As shown in the middle sections of Table~\ref{tab:main_table} and~\ref{tab:hpftable}, our weakly-supervised model also achieves the state of the art in the weakly-supervised regime.  
In particular, our model shows more reliable transferablility compared to strongly-supervised models, outperforming both weakly~\cite{huang2019dynamic} and strongly-supervised~\cite{min2019hyperpixel} state of the arts by 6.4\% and 5.8\% points of PCK respectively on PF-WILLOW. On the Caltech-101 benchmark, our method is comparable to the best among the recent methods. 
Note that unlike other benchmarks, the evaluation metric of Caltech-101 is indirect (\ie, accuracy of mask transfer). 
On the SPair-71k dataset, where image pairs have large view point and scale differences, the methods of~\cite{rocco18weak,rocco2018neighbourhood} as well as ours do not successfully learn in the weakly-supervised regime; they (FT) all underperform transferred models (TR) trained on PF-PASCAL. This result reveals current weakly-supervised objectives are all prone to large variations, which requires further research in the future.

\smallbreak
\noindent \textbf{Effect of layer selection rate $\mu$~\cite{veit2018convolutional}.}
The plot in Fig.~\ref{fig:layer_selection}a shows PCK and running time of our models trained with different layer selection rates $\mu$. It shows that smaller selection rates in training lead to faster running time in testing, at the cost of some accuracy, by encouraging the model to select a smaller number of layers. The selection rate $\mu$ can thus be used for speed-accuracy trade-off.  

\begin{figure}[!t]
  \centering
  \begin{minipage}[b]{0.58\textwidth}
    \includegraphics[width=\textwidth]{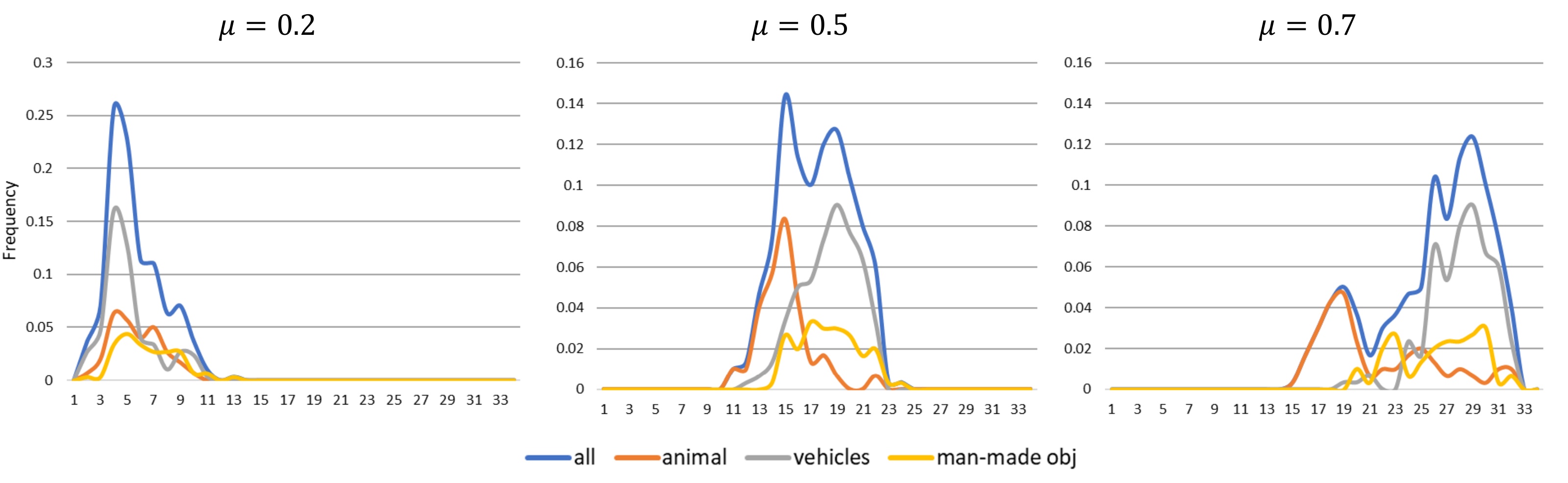}
    \caption{\label{fig:num_of_layer_used_histo}Frequencies over the numbers of selected layers with different selection rates $\mu$ (x-axis: the number of selected layers, y-axis: frequency). Best viewed in electronics.}
  \end{minipage}
  \hfill
  \begin{minipage}[b]{0.4\textwidth}
    \includegraphics[width=\textwidth]{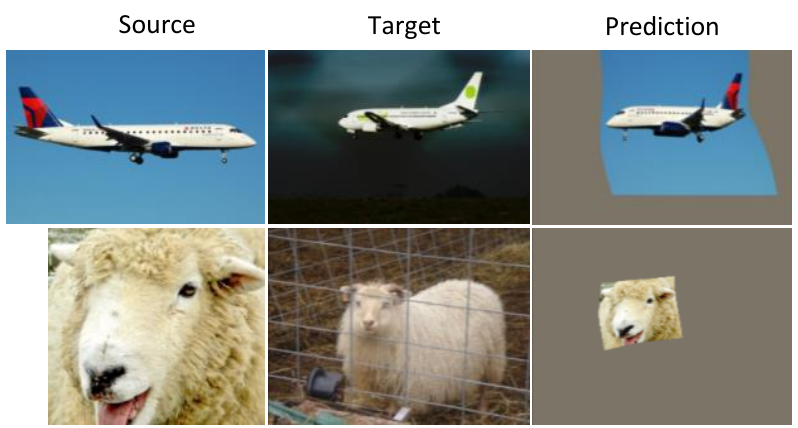}
    \caption{\label{fig:qualitative_spair}Example results on SPair-71k dataset. The source images are warped to the target ones using resultant correspondences.}
  \end{minipage}
\end{figure}

\begin{figure}[t]
    \centering
    \scalebox{0.6}{
    \includegraphics{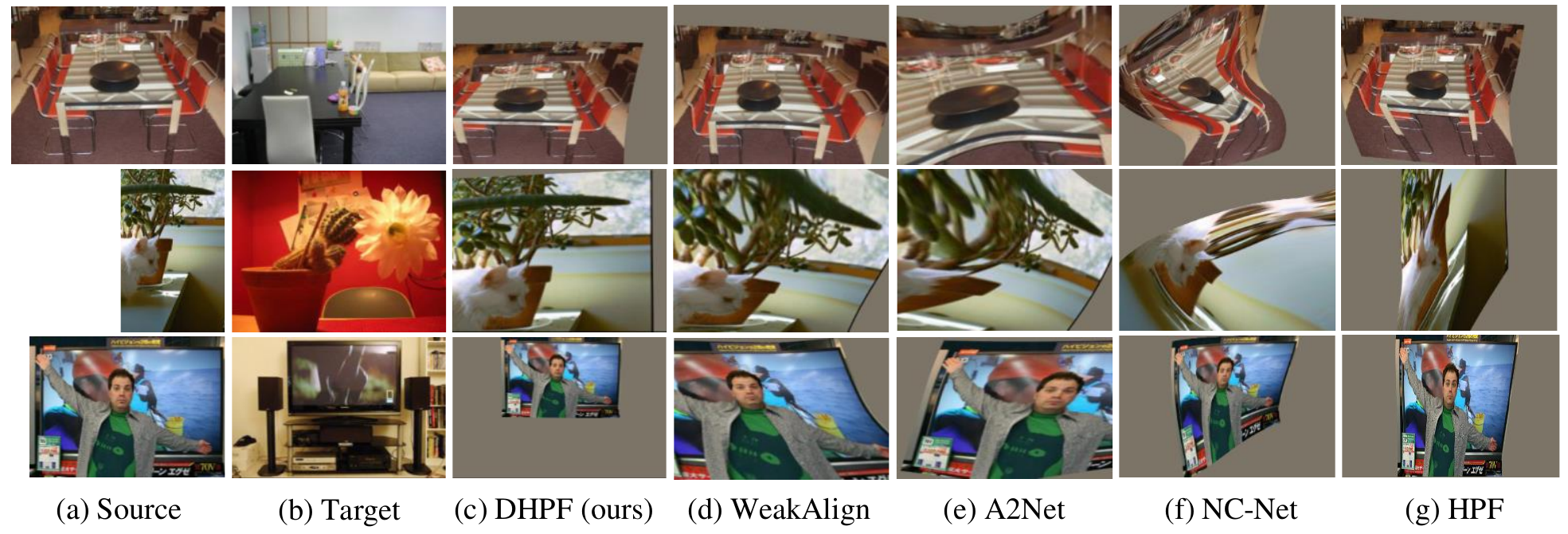}
    }
    \caption{Example results on PF-PASCAL~\cite{ham2018proposal}: (a) source image, (b) target image and (c) DHPF (ours), (d) WeakAlign~\cite{rocco18weak}, (e) A2Net~\cite{paul2018attentive}, (f) NC-Net~\cite{rocco2018neighbourhood}, and (g) HPF~\cite{min2019hyperpixel}.}
    \label{fig:vis_pfpascal}
\end{figure}

\smallbreak
\noindent \textbf{Analysis of layer selection patterns.}
Category-wise layer selection patterns in Fig.~\ref{fig:layer_selection}b show that each group of animal, vehicle, and man-made object categories shares its own distinct selection patterns.
The model with a small rate ($\mu=0.3$) tends to select the most relevant layers only while the model with larger rates ($\mu>0.3$) tends to select more complementary layers as seen in Fig.\ref{fig:layer_selection}c.
For each $\mu \in \{0.3, 0.4, 0.5\}$ in Fig.\ref{fig:layer_selection}c, the network tends to select low-level features for vehicle and man-made object categories while it selects mostly high-level features for animal category.
We conjecture that it is because low-level (geometric) features such as lines, corners and circles appear more often in the vehicle and man-made classes compared to the animal classes.
Figure~\ref{fig:num_of_layer_used_histo} plots the frequencies over the numbers of selected layers with different selection rate $\mu$, where vehicles tend to require more layers than animals and man-made objects. 

\smallbreak
\noindent \textbf{Qualitative results.}
Some challenging examples on SPair-71k~\cite{min2019spair} and PF-PASCAL~\cite{ham2018proposal} are shown in Fig.\ref{fig:qualitative_spair} and \ref{fig:vis_pfpascal} respectively: Using the keypoint correspondences, TPS transformation~\cite{donato2002approximate} is applied to source image to align target image.
The object categories of the pairs in Fig.\ref{fig:vis_pfpascal} are in order of table, potted plant, and tv.
Alignment results of each pair demonstrate the robustness of our model against major challenges in semantic correspondences such as large changes in view-point and scale, occlusion, background clutters, and intra-class variation.

\smallbreak
\noindent \textbf{Ablation study.}
We also conduct an ablation study to see the impacts of major components: Gumbel layer gating (GLG), conv feature transformation (CFT), probabilistic Hough matching (PHM), keypoint importance weight $\omega_m$, and layer selection loss $\mathcal{L}_{\mathrm{sel}}$.
All the models are trained with strong supervision and evaluated on PF-PASCAL. 
Since the models with a PHM component have no training parameters, they are directly evaluated on the test split.
Table~\ref{tab:ablation_study} summarizes the results. 
It reveals that among others CFT in the dynamic gating module is the most significant component in boosting performance and speed; without the feature transformation along with channel reduction, our models do not successfully learn in our experiments and even fail to achieve faster per-pair inference time.
The result of `w/o $\omega_m$' reveals the effect of the keypoint weight $\omega_m$ in  Eq.(\ref{one_hot_gce}) by replacing it with uniform weights for all $m$, \ie, $\omega_m = 1$; putting less weights on easy examples helps in training the model by focusing on hard examples.
The result of `w/o $\mathcal{L}_{\mathrm{sel}}$' shows the performance of the model using $\mathcal{L}_{\mathrm{match}}$ only in training; performance drops with slower running time, demonstrating the effectiveness of the layer selection constraint in terms of both speed and accuracy.
With all the components jointly used, our model achieves the highest PCK measure of $90.7\%$. Even with the smaller backbone network, ResNet-50, the model still outperforms previous state of the art and achieves real-time matching as well as described in Fig.\ref{fig:layer_selection} and Table \ref{tab:main_table}. 

\begin{table}[!t]
    \begin{minipage}{.51\linewidth}
        \caption{\label{tab:ablation_study}Ablation study on PF-PASCAL. (GLG: Gumbel layer gating with selection rates $\mu$, CFT: conv feature transformation)}
        \begin{center}
            \begin{tabular}{ccccccc}
                
                    \toprule
                    \multicolumn{3}{c}{Module} & \multicolumn{3}{c}{PCK ($\alpha_{\text{img}}$)} & time \\
                    GLG & CFT & PHM  & $0.05$ & $0.1$ & $0.15$ & ({\em ms}) \\ 
                    \midrule
                    0.5 & \cmark & \cmark & 75.7 & 90.7 & 95.0 & 58 \\
                    0.4 & \cmark & \cmark & 73.6 & 90.4 & 95.3 & 51 \\
                    0.3 & \cmark & \cmark & 73.1 & 88.7 & 94.4 & 47 \\
                    \midrule
                    & \cmark & \cmark & 70.4 & 88.1 & 94.1 & 64 \\
                    0.5 &  & \cmark & 43.6 & 74.7 & 87.5 & 176 \\ 
                    0.5 & \cmark & & 68.3 & 86.9 & 91.6 & 57 \\
                    &  & \cmark & 37.6 & 68.7 & 84.6 & 124 \\
                    & \cmark &  & 68.1 & 85.5 & 91.6 & 61 \\
                    0.5 &  &  & 35.0 & 54.8 & 63.4 & 173 \\
                    \midrule
                    \multicolumn{3}{c}{ w/o  $\omega_m$ } & 69.8 & 86.1 & 91.9 & 57 \\
                    \multicolumn{3}{c}{ w/o  $\mathcal{L}_{\mathrm{sel}}$} & 68.1 & 89.2 & 93.5 & 56 \\
                    \bottomrule
            \end{tabular}
        \end{center}
    \end{minipage}
    \begin{minipage}{.48\linewidth}
        \caption{\label{tab:ablation_study_gating}Comparison to soft layer gating on PF-PASCAL.}

        \begin{center}
            \begin{tabular}{ccccc}
                \toprule
                \multirow{2}{*}{Gating function} & \multicolumn{3}{c}{PCK ($\alpha_{\text{img}}$)} & time \\
                & $0.05$ & $0.1$ & $0.15$ & ({\em ms}) \\ 
                \midrule
                Gumbel$_{\mu=0.5}$  & 75.7 & 90.7 & 95.0 & 58 \\
                \midrule
                sigmoid &  71.1 & 88.2 & 92.8 & 74 \\
                sigmoid$_{\mu=0.5}$ & 72.1 & 87.8 & 93.3 & 75 \\
                sigmoid + $\ell1$& 65.9 & 87.2 & 91.0 & 60 \\
                \bottomrule
        \end{tabular}
        \end{center}
        
        \begin{center}
        \includegraphics[width=1.0\linewidth]{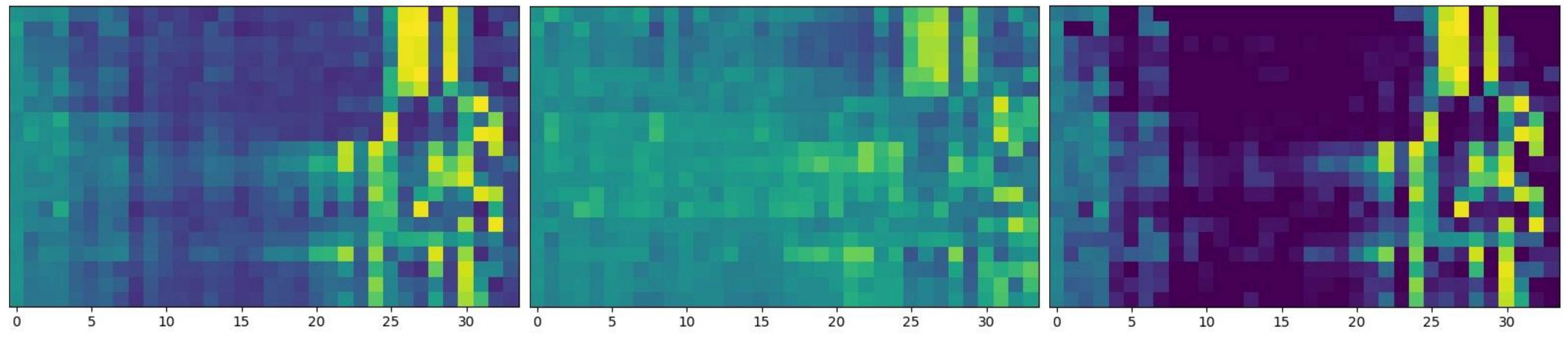}
        \end{center}
        
    	\captionof{figure}{\label{fig:vis_layer_soft_sparse} ResNet-101 layer selection frequencies for `sigmoid' (left), `sigmoid$_{\mu=0.5}$' (middle), and `sigmoid + $\ell1$' (right) gating.}
    \end{minipage} 
\end{table}

\smallbreak
\noindent \textbf{Computational complexity.}
The average feature dimensions of our model before correlation computation are 2089, 3080, and 3962 for each $\mu \in \{0.3, 0.4, 0.5\}$ while those of recent methods \cite{min2019hyperpixel,lee2019sfnet,rocco2018neighbourhood,huang2019dynamic} are respectively 6400, 3072, 1024, 1024. The dimension of hyperimage is relatively small as GLG efficiently prunes irrelevant features and CFT effectively maps features onto smaller subspace, thus being more practical in terms of speed and accuracy as demonstrated in Table~\ref{tab:main_table} and \ref{tab:ablation_study}. Although \cite{rocco2018neighbourhood,huang2019dynamic} use lighter feature maps compared to ours, a series of 4D convolutions heavily increases time and memory complexity of the network, making them expensive for practical use (31ms (ours) vs. 261ms \cite{rocco2018neighbourhood,huang2019dynamic}).

\subsection{Comparison to soft layer gating} \label{sec:softgating}
The Gumbel gating function in our dynamic layer gating can be replaced with conventional soft gating using sigmoid.  
We have investigated different types of soft gating as follows:   
(1) `sigmoid': The MLP of dynamic gating at each layer predicts a scalar input for sigmoid and the transformed feature block pairs are weighted by the sigmoid output.
(2) `sigmoid$_{\mu=0.5}$': In training the `sigmoid' gating, the layer selection loss $\mathcal{L}_\mathrm{sel}$ with $\mu=0.5$ is used to encourage the model to increase diversity in layer selection.
(3) `sigmoid + $\ell1$': In training the `sigmoid' gating, the $\ell1$ regularization on the sigmoid output is used to encourage the soft selection result to be sparse. 
Table~\ref{tab:ablation_study_gating} summarizes the results and Fig. \ref{fig:vis_layer_soft_sparse} compares their layer selection frequencies.  

While the soft gating modules provide decent results, all of them perform worse than the proposed Gumbel layer gating in both accuracy and speed. 
The slower per-pair inference time of `sigmoid' and `sigmoid$_{\mu=0.5}$' indicates that {\em soft} gating is not effective in skipping layers due to its non-zero gating values. We find that the sparse regularization of `sigmoid + $\ell1$' recovers the speed but only at the cost of significant accuracy points.  
Performance drop of soft gating in accuracy may result from the {\em deterministic} behavior of the soft gating during training that prohibits exploring diverse combinations of features at different levels. 
In contrast, the Gumbel gating during training enables the network to perform more comprehensive trials of a large number of different combinations of multi-level features, which help to learn better gating.
Our experiments also show that {\em discrete} layer selection along with {\em stochastic} learning in searching the best combination is highly effective for learning to establish robust correspondences in terms of both accuracy and speed.

% !TEX root = ../main.tex

%%%%%%%%% 5. CONCLUSION
\section{Conclusion}
We have presented a dynamic matching network that predicts dense correspondences by composing hypercolumn features using a small set of relevant layers from a CNN. 
The state-of-the-art performance of the proposed method indicates that the use of dynamic multi-layer features in a trainable architecture is crucial for robust visual correspondence. 
We believe that our approach may prove useful for other domains involving correspondence such as image retrieval, object tracking, and action recognition.
We leave this to future work.

\bigskip
\noindent \textbf{Acknowledgements.}
This work is supported by Samsung Advanced Institute of Technology (SAIT) and also by Basic Science Research Program (NRF-2017R1E1A1A01077999) and Next-Generation Information Computing Development Program (NRF-2017M3C4A7069369) through the National Research Foundation of Korea (NRF) funded by the Ministry of Science, ICT, Korea. 
Jean Ponce was supported in part by the Louis Vuitton/ENS chair in artificial intelligence and the Inria/NYU collaboration and also by the French government under management of Agence Nationale de la Recherche as part of the "Investissements dâavenir" program, reference ANR-19-P3IA-0001 (PRAIRIE 3IA Institute). 

{\small
    \bibliographystyle{splncs04}
    \bibliography{egbib}
}
\end{document}